\documentclass[5p,twocolumn,10pt,times]{elsarticle}
\usepackage{amsmath}
\usepackage{algorithm}
\usepackage{algpseudocode}
\usepackage{hyperref}
\usepackage{wrapfig}
\usepackage{cuted}
\usepackage{capt-of}
\begin{document}
\baselineskip11pt

\begin{frontmatter}

\title{Self-Parameterization Based Multi-Resolution Mesh Convolution Networks}

\author{Shi Hezi}
\author{Jiang Luo}
\cortext[mycorresponingauthor]{Corresponding author}
\ead{luo.jiang@ntu.edu.sg}
\author{Zheng Jianmin}
\address{HP-NTU Digital Manufacturing Corporate Laboratory, Nanyang Technological University}
\author{Zeng Jun}
\address{HP Labs}

\begin{abstract} This paper addresses the challenges of designing mesh convolution neural networks for 3D mesh dense prediction. While deep learning has achieved remarkable success in image dense prediction tasks, directly applying or extending  these methods to irregular graph data, such as 3D surface meshes, is nontrivial due to the non-uniform element distribution and irregular connectivity in surface meshes which make it difficult to adapt downsampling, upsampling, and convolution operations. In addition, commonly used multiresolution networks require repeated high-to-low and then low-to-high processes to boost the performance of recovering rich, high-resolution representations. 
To address these challenges, this paper proposes a self-parameterization-based multi-resolution convolution network that extends existing image dense prediction architectures to 3D meshes. The novelty of our approach lies in two key aspects.
First, we construct a multi-resolution mesh pyramid directly from the high-resolution input data and propose area-aware mesh downsampling/upsampling operations that use sequential bijective inter-surface mappings between different mesh resolutions. The inter-surface mapping redefines the mesh, rather than reshaping it, which thus avoids introducing unnecessary errors. Second, we maintain the high-resolution representation in the multi-resolution convolution network, enabling multi-scale fusions to exchange information across parallel multi-resolution subnetworks, rather than through connections of high-to-low resolution subnetworks in series. These features differentiate our approach from most existing mesh convolution networks and enable more accurate mesh dense predictions, which is confirmed in experiments.
\end{abstract}

\end{frontmatter}

%\linenumbers

\section{Introduction}

\begin{figure*}
\centering
\includegraphics[width=0.95\textwidth]{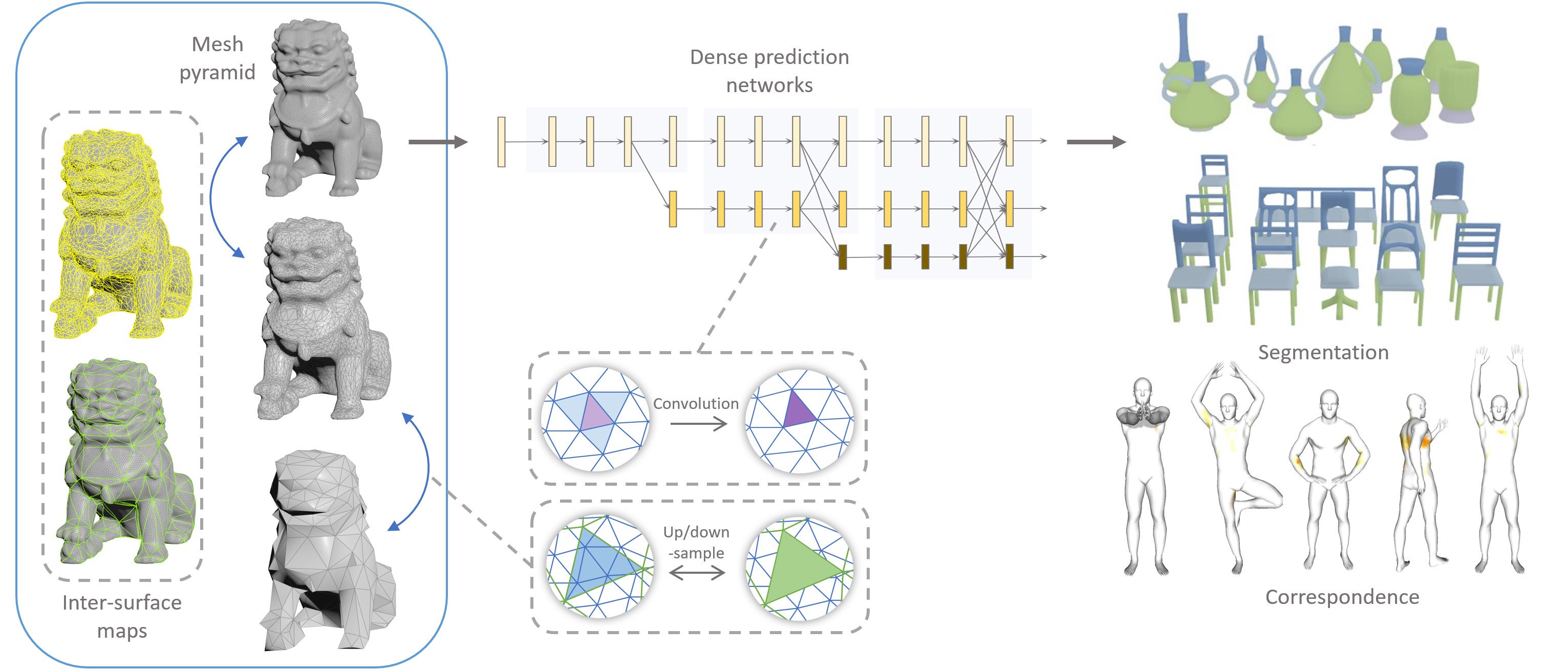}
\caption{\label{fig:teasure} We construct
 a multi-resolution mesh pyramid from high-resolution input meshes using bijective surface-to-surface maps for any pair of adjacent meshes in the pyramid (left). Based on this, we design down-/up-sampling operators that enable us to easily adapt classic CNN dense prediction architectures and we construct a multi-resolution convolution network that maintains high-resolution representations throughout the network (middle). The resulting networks allow for both high-to-low and low-to-high feature aggregations, leading to high performance in mesh dense predictions (right).}
 \label{fig:teaser}
\end{figure*}

This paper considers 3D mesh dense prediction tasks such as instance segmentation, semantic segmentation, monocular depth estimation, human pose estimation, and shape correspondence, which are often required in geometric modeling and analysis. In the past decade, deep learning has achieved remarkable success in various 2D/3D applications, including dense predictions in computer graphics and vision. Some well-known networks such as AlexNet~\cite{alexnet2012}, VGG~\cite{VGG2015} and ResNet~\cite{resnet2016} have been developed. However, dense predictions are position-sensitive and often require element-level predictions, making classical network architectures that first encode the input data into low-resolution representations  unsuitable for these tasks. Therefore, new architectures and strategies are needed to achieve accurate dense predictions.

There are typically three approaches to generating high-resolution representations for images using convolutional neural networks (CNN). The first approach involves a high-to-low and low-to-high process, which recovers high-resolution features through upsampling operations~\cite{deconvnet2015,segnet2017} and skip-connections~\cite{unet2015,hourglass2016}. The second approach is to build an image pyramid with pyramid pooling modules~\cite{PSPNet2017} or atrous spatial pyramid pooling~\cite{deeplabv32016,deeplabv3plus2018}. The pyramid features are then fused to generate high-resolution representations. The third approach maintains high-resolution channels through the high-to-low and low-to-high process, together with repeated multi-scale feature fusion modules, which produces representations with rich semantic meaning~\cite{HRnet1,HRnet2} and has been reported to achieve state-of-the-art dense prediction performance.

%This paper considers 3D mesh dense prediction tasks such as instance segmentation, semantic segmentation, monocular depth estimation, human pose estimation, and shape correspondence, which are often required in geometric modeling and analysis. In the past decade, deep learning has achieved remarkable success in various image applications, including dense predictions in computer graphics and vision. Dense predictions are position-sensitive and often require element-level predictions, which are more challenging than recognition and retrieval tasks. There is a lot of work focusing on dense prediction of images, The third approach maintains high-resolution channels through the high-to-low and low-to-high process, together with repeated multi-scale feature fusion modules, which produces representations with rich semantic meaning~\cite{HRnet1,HRnet2} and has been reported to achieve state-of-the-art dense prediction performance.

Applying and extending CNN frameworks for 3D mesh dense prediction is a tempting idea, but the irregular structure of 3D meshes presents a challenge. Basic CNN operators, such as convolution, pooling, and unpooling, need to be specifically designed to work with non-grid structures. 
Recently, some approaches have been proposed to mimic CNN networks, such as MeshCNN~\cite{MeshCNN2019} and PDMeshNet~\cite{PDMeshNet2020}, which utilize sequential edge collapses to produce a lower-resolution mesh. However, the selection of edges for collapse depends on specific learning tasks, and training samples may have specific regions that never collapse. This may raise the requirement for diversity of training data and increase the complexity of training networks. Another approach, 
SubdivNet~\cite{hu2021subdivision}, proposes to remesh the irregular input mesh into a semi-regular structure with subdivision sequence connectivity. However, this approach applies the network operators to the remeshed model instead of the original mesh, making the method dependent on the remeshing results.

Thus one question arises: {\em can we construct a neural network defined directly on irregular mesh data whose local receptive fields can grow uniformly after each convolution, pooling, or unpooling layer}? Though the terms ``irregularity'' and ``uniformity'' seem contradictory, this paper provides a positive answer to this question. In particular, to achieve top-down and bottom-up propagation, we first create a multi-resolution mesh using mesh simplification~\cite{qem97}. To address the challenge of performing basic network operations on the irregular connectivity of a triangular mesh, we construct bijective piecewise linear surface-to-surface maps between meshes of adjacent levels, inspired by inter-surface mapping~\cite{intersurface2004}. Specifically, we map each triangle of a coarse mesh to a ``curved'' triangle on the fine mesh. The curved triangle consists of polygonal pieces on the fine mesh, which we can use to design area-aware pooling/unpooling and face convolution layers. For example, in the pooling layer, we merge all the polygonal pieces associated with a curved triangle into one, and the contribution of each polygonal piece is proportional to its area.
The key to our approach is that the inter-surface mapping \emph{redefines} the input mesh rather than \emph{reshaping} it. This provides guidance for feature propagation without introducing approximation errors or worrying about the irregularity of the input data. 
Furthermore, we control the triangle quality during the construction of the multi-resolution mesh, as described in~\cite{qem97,global_para}, to ensure that the triangles in each level have similar shapes. 
Additionally, we can constrain the surface-to-surface maps to minimize significant distortion. As a result, the local receptive fields of our network have a relatively fixed pattern and expand uniformly during the layer-by-layer pass, similar to powerful 2D CNNs. Finally, we propose to maintain the high-resolution representation in the  multi-resolution convolution network, which enables multi-scale fusion to exchange information across parallel multi-resolution subnetworks and avoids loss of information in the high-to-low process. 
The overview of our approach is illustrated in Figure.~\ref{fig:teaser}.

The contributions of the paper are three-folds. First, we propose a new method for constructing a multi-resolution mesh pyramid associated with piecewise bijective surface mapping among adjacent levels of the mesh. Second, we design novel area-aware pooling/unpooling and face convolution operations that leverage the bijective surface-to-surface mapping to achieve efficient mesh feature learning. Third, by utilizing our multi-resolution architectures and novel operators, we introduce a new framework called {\em Self-Parameterization Based Multi-Resolution Mesh Convolution Networks} (SPMM-Net), built on existing CNN architecture HRNet~\cite{HRnet1,HRnet2}, to tackle mesh dense prediction tasks. The experiments demonstrate that our proposed method achieves better performance than the state-of-the-art.

\section{Related Work}

This section briefly reviews some related work, which includes geometric deep learning, especially those adapting classic image dense prediction architectures to mesh structures, and inter-surface mapping.  

%-------------------------------------------------------------------------
\subsection{Geometric Deep Learning for 3D Shape}
A surface mesh is composed of vertices, edges, and faces. 
Geometric deep learning methods perform convolution in local Euclidean neighborhoods based on the connectivity of vertices or the topology of surfaces. 
 Various convolution operations have been designed.  The methods that perform  convolution on vertices consider the spatial relations of vertices using Cartesian, polar or spherical coordinates, and define weights between vertices accordingly. 
Early approaches focused on adapting  grid CNNs to irregular graph data. GCNN~\cite{GCNN2015} proposed building a local geodesic system in polar coordinates for each node and discretizing neighboring nodes into a fixed number of bins to handle nodes without a fixed number of neighbors. MoNet~\cite{MoNet2017} represented the learnable convolution kernel in 3D space using a Gaussian Mixture Model (GMM), while SplineCNN~\cite{SplineCNN2018} extended MoNet with a more flexible convolution kernel that uses B-spline basis functions in a geometric space, providing more learnable parameters compared to previous methods. 

Face-based methods incorporate topological features into mesh faces and use an aggregation mechanism to combine neighboring face features. MeshNet~\cite{feng2019meshnet} introduced a face rotation convolution block that applies the convolution kernel to three pairs of corner vectors, combining the results with MLP layers to ensure invariance to the order of corner vectors. Another method~\cite{Schult2020dual} defines two types of convolutions based on geodesic and Euclidean distances to improve the convolution results. SubdivNet~\cite{hu2021subdivision} remeshes the input mesh into Loop subdivision connectivity to enable four adjacent faces to be pooled into one face. It also introduces an order-invariant face convolution and a mesh pooling operation. Instead of using the geometric descriptor, Laplacian Mesh Transformer~\cite{meshtransformer2022} uses Laplacian spectral decomposition to describe the connectivity of triangles, and employs dual attention mechanism to define the message propagation process.

In edge-based convolution, MeshCNN~\cite{MeshCNN2019} defines convolution operators directly on edges, leveraging the consistent neighboring structure of edges in a triangular mesh.  MeshWalker~\cite{lahav2020meshwalker} extracts shape features by walking randomly along mesh edges to sample subgraphs for learning. PD-MeshNet~\cite{PDMeshNet2020} constructs both a primal and a dual graph, enabling the assignment of features to edges and faces as well.

For dense prediction tasks on meshes, adapting existing 2D CNN techniques, such as high-resolution representation recovery and multi-resolution fusion, is an appealing idea. However, defining multi-scale representations on meshes is non-trivial due to their irregular and non-uniform characteristics. Clustering methods, such as Graclus pooling~\cite{Defferrard2016} (also called greedy node clustering) and diff-pool~\cite{diffpool2018}, have been proposed for graph pooling. Although these techniques work well for some undirected graphs, they may not preserve the manifold structure, resulting in undesirable outcomes.
One possible solution is to re-design the pooling operation, for instance, using dynamic edge collapse operations which generate a coarse surface, as demonstrated in MeshCNN~\cite{MeshCNN2019} and PD-MeshNet~\cite{PDMeshNet2020}. These methods record the indices and positions of the collapsed edges, which can be used to perform upsampling via vertex split operations and thus to adapt CNN dense prediction networks such as U-Net. However, they may result in non-uniform downsampling of edges. Another solution is to build multi-resolution mesh pyramids through remeshing, as shown in SubdivNet~\cite{hu2021subdivision} and Neural Subdivision~\cite{liu2020neural}. These methods first decimate the input mesh to a base mesh and then construct a semi-regular connectivity by subdividing the base mesh. The semi-regular structure allows for the natural extension of operators in CNN. However, these methods heavily rely on the quality of the remeshing results.

%%%%%%%%%%%%%%
%Various symmetric networks~\cite{segnet2017,unet2015,hourglass2016,deconvnet2015} first downsample the input image, then perform upsampling to recover the resolution, and skip the connection between the same-resolution layers. However, the downsampling process reduces the sensitivity of position, which may lower the accuracy of pixel-level predictions. To overcome this issue, some methods were proposed, which maintain the high-resolution feature through layers~\cite{msdensenet2017,HRnet1}. These methods consist of multiple resolution channels and inter-channel information exchange modules. Input image are passed on different resolution channels with accurate position information on high-resolution channels and wide receptive fields on low-resolution channels. The inter-channel information is merged with operations including pooling, strided convolution, deconvolution and upsampling. Building  image/feature pyramids can generate multi-resolution representations. The SPPNet~\cite{SPPnet} builds feature pyramids after convolution layers. The feature maps are pooled with different stride pooling operations and then concatenated as the input to the next layer. Our work will show that with properly designed operators on irregular meshes, some of existing CNN architectures can be easily adapted to maintain multi-resolutions in parallel, which can substantially improve 3D mesh dense prediction performance.

%%%%%%%%%%%%%%
\subsection{Inter-surface Mapping}
Inter-surface mapping is to create a continuous and bijective surface-to-surface map between two triangle meshes with the same topology. Existing methods can be roughly divided into two categories according to whether they need some intermediate domains or not. The first category employs a unit disk or sphere as the intermediate domain, with each triangular mesh parameterized on this domain. The inter-surface map is then defined as the composite of separate parameterizations~\cite{lifted2014,disk_para2019}. As
these methods cannot be directly applied to surfaces with different topologies, cut/transition processes are required~\cite{orbifold2015,sphereTutte2017}, which adds extra steps. 
Recently, constant-curvature metrics have been employed to optimize the end-to-end map without cuts, providing a solution for surfaces with arbitrary genus~\cite{intersurface_curv2020}. 
The second category uses local optimization to generate a continuous map between discrete surfaces with arbitrary topology~\cite{intersurface2004,kraevoy2004cross,hsuehti2021}. However, this approach is difficult to generate globally optimal maps and also requires either solving a complicated  optimization problem or introducing additional processing such as designing cuts, making it time-consuming and difficult to handle large-scale and multi-class datasets.

%------------------------------------------------------------------------
%%%%%%%%%%%%%%%%%%%%%%%%%%%%%%%%%%%%%%%%%%%%%%%%%%%%%
\section{Mesh Pyramid}
In our proposed framework, the input mesh must undergo a process to create a mesh pyramid consisting of multiple resolutions from fine to coarse, and a set of bijective surface-to-surface maps before it is fed into the network. This section presents an enhanced self-parameterization based approach for generating a mesh pyramid and  bijective maps between the adjacent levels of the multi-resolution meshes.

\begin{figure}[htb]
  \centering
  \includegraphics[width=1.0\linewidth]{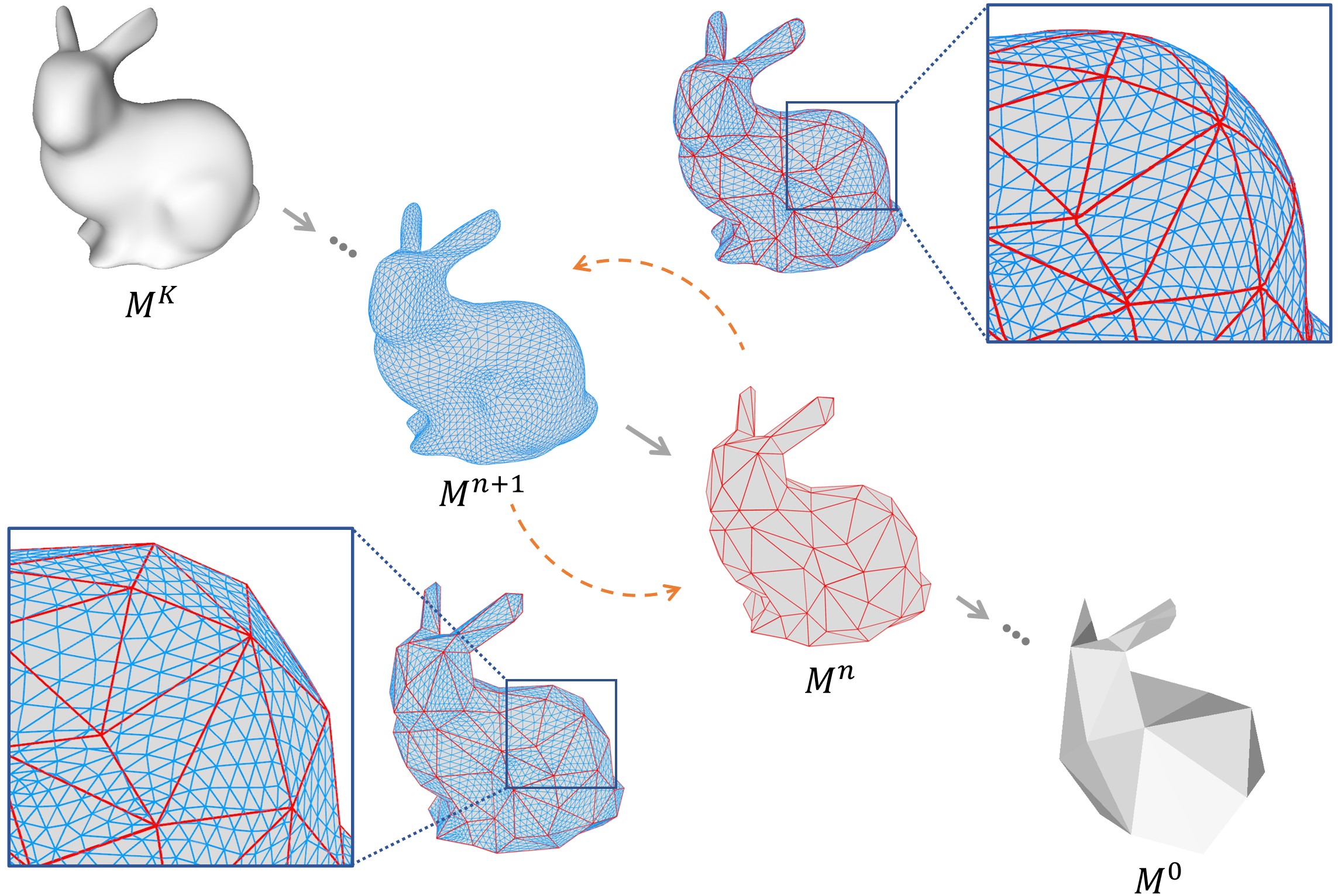}
  \parbox[t]{\columnwidth}{\relax}
  \caption{\label{fig:mesh_sequence}
            Generating the multi-resolution mesh pyramid from an input mesh. Bijective inter-surface maps between $M^{n+1}$ (blue) and $M^{n}$ (red) are visualizaed.}
\end{figure}

%%%%%%%%%%%%%%%%%%%%%%%%
\subsection{Notation}
The input triangle mesh is 2-manifold and watertight. We denote it by $M=\left(V, E, F\right)$, where $V=\{v_{i}\mid v_{i}\in\mathbb{R}^{3}\}$, $E=\{e_i\mid e_i = (v_{i_1}, v_{i_2}\}$ and $F=\{f_{i}\mid f_{i} = (e_{i_1},e_{i_2}, e_{i_3})\}$ are a set of vertices, a set of edges and a set of triangular faces, respectively. Each edge is bounded by two vertices, and each face is bounded by three edges. Two faces are adjacent if they share an edge, and each face $f_{i}$ has three adjacent faces $\{f_{i_j}\mid j = 1,2,3\}$.

With an input mesh $M$, we define a sequence of decimated meshes $(M^{K},\cdots,M^{0})$ such that $M^{K}=M$, and $M^{0}$ is the base mesh. $K+1$ is the height of the mesh pyramid we want to create. Moreover, a bijective surface-to-surface map $\mathcal{H}^n$ from $M^{n+1}$ to $M^{n}$ is also defined, for all $n \in \{K-1, K-2, \dots, 0\}$. It is worth pointing out that our $\mathcal{H}^n(\cdot)$ is defined for every point on the mesh $M^{n+1}$, not just its vertices.

In addition, a sparse matrix $A^{n}=\left[a^n_{i,j}\right]\in \mathbb{R}^{|F^{n+1}|\times|F^n|}$ is computed  using the bijective inter-surface map $\mathcal{H}^n: M^{n+1} \rightarrow M^{n}$, $0 \le n \le K-1 $, where $a^n_{i,j} \in [0, 1]$ measures the overlap ratio between  face $f^{n}_j$ of mesh $M^{n}$ and image $\mathcal{H}^{n}(f^{n+1}_{i})$ of face $f^{n+1}_i$ of mesh $M^{n+1}$. The sparse matrices are used to define the downsampling and upsampling operations between two adjacent pyramid levels.

\subsection{Multi-Resolution Mesh Generation}

Given a high-resolution input mesh $M$, our target is to construct a set of multi-resolution meshes $\{M^{n}\}_{n=0}^K$ and a set of bijective inter-surface maps $\{\mathcal{H}_n: M^{n+1} \rightarrow M^{n}\}_{n=0}^{K-1}$ (see Figure.~\ref{fig:mesh_sequence}) such that the triangles of each mesh except $M^{K}$ (the input one) have a similar shape and all the surface maps have small distortions. Note that such requirements imposed on the triangles and the maps can make our network defined on irregular data more similar to traditional CNN. %Moreover, the inter-surface map with large distortion could weaken the role of local features during aggregation and suppress the performance of dense prediction tasks.

% One possible solution is to use the MAPS algorithm~\cite{maps1998} that can generate hierarchical mesh simplification and establish a vertex-to-surface map between any pair of adjacent mesh levels. Unfortunately, none of these methods can produce a whole surface-to-surface map. An easy remedy is to use recently developed inter-surface map techniques~\cite{intersurface2004}\cite{intersurface_curv2020} to build such maps with very low distortion for any pair of neighbouring mesh levels. However, these methods often need to solve an expensive optimization problem, which can have high computation time and be sensitive to user-specified parameters. So they are impractical for us to construct a large-scale training data.

\begin{figure}[htb]
\centering
\includegraphics[width=0.9\columnwidth]{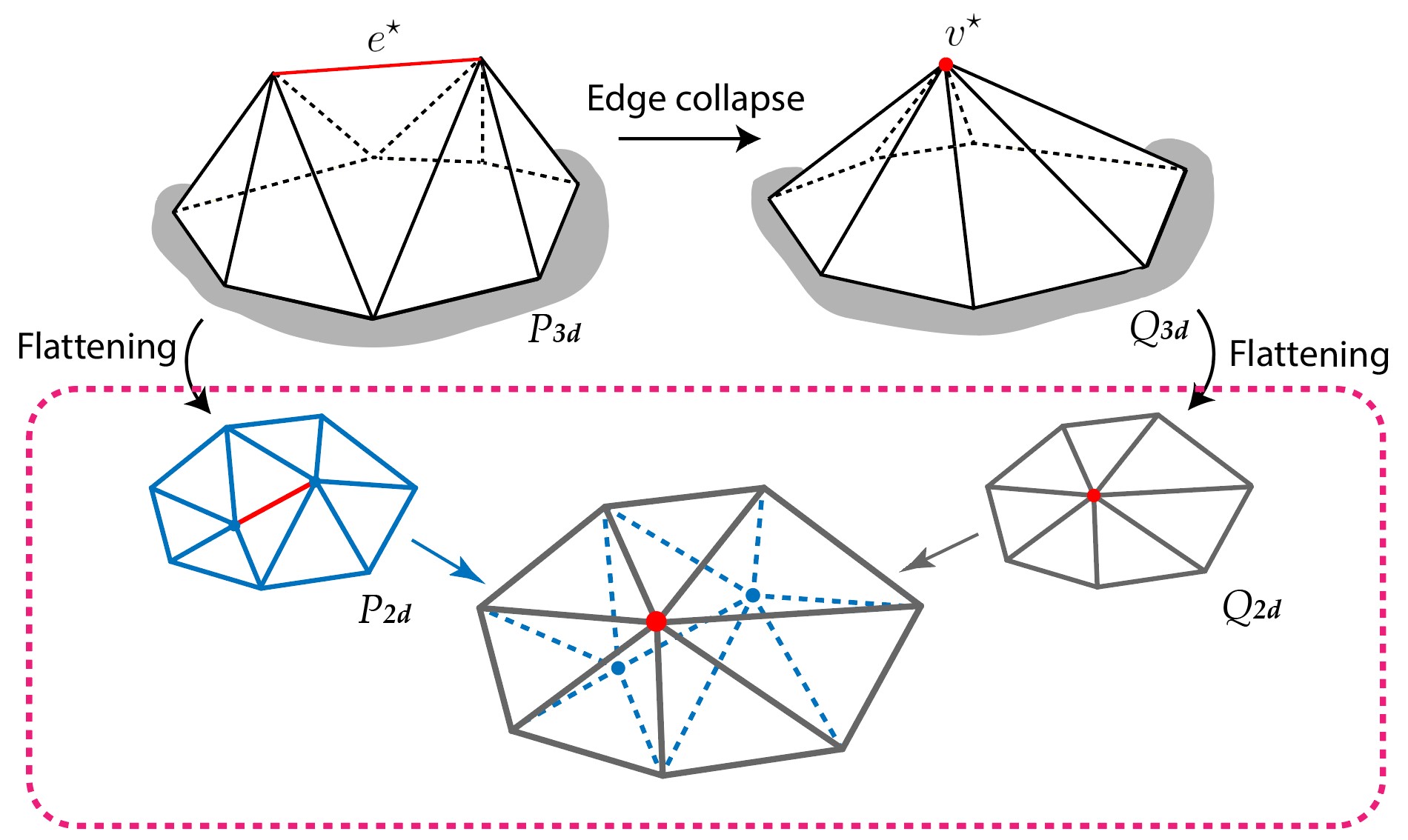}
\caption{We flatten the one-ring face-neighborhood of $e^{\star}$ using LSCM parameterization~\cite{lscm2002}. After $v^{\star}$ is generated by QEM in 3D space, we also flatten its one-ring face-neighborhood into the same UV domain. \label{fig:initail_collapse}}
\end{figure}

\begin{figure*}
\centering
\includegraphics[width=0.99\textwidth]{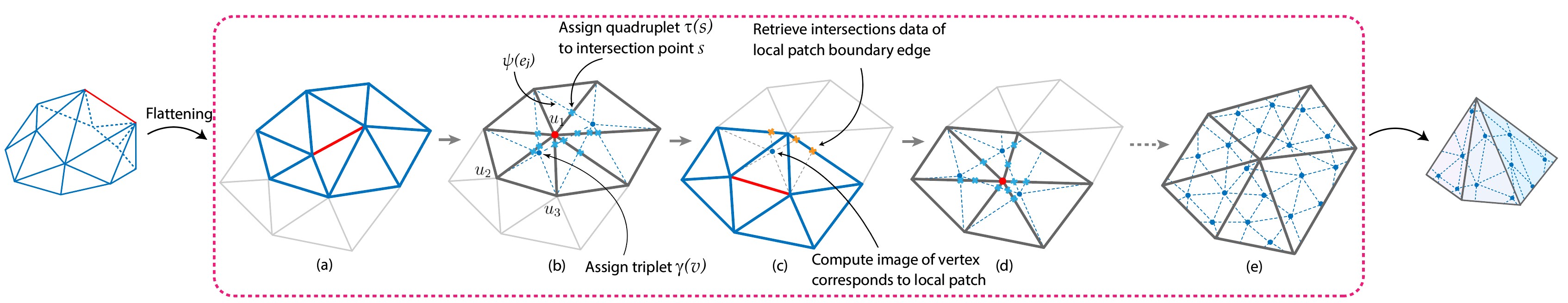}
\caption{\label{fig:mapping} The process of recording and retrieving edge-to-edge intersection information during two consecutive iterations. From left to right: (a) A 3D patch is flattened and the edge $e^{\star}$ is to be collapsed. (b) The intersection points are computed in the UV domain and recorded as quadruplets, while the collapsed vertices are recorded as triplets. (c) In the second iteration, intersection points on the boundary edges of the blue patch are retrieved according to recorded quadruplets, and the triplets corresponding to the blue patch are also updated. (d) Intersection points for inner edges of the local patch are updated. (e) After several iterations, each face of the local patch contains a partial map of $M^{n+1}$. }
\end{figure*}

Here we present a simple and efficient approach that is based on MAPS~\cite{maps1998}. The mesh pyramid can be generated through various implementations, which can either rely on vertex removal or edge collapse. We opt for an implementation~\cite{liu2020neural} that utilizes QEM~\cite{qem97} to steer edge collapse in a more balanced way, while disregarding collapses that could result in triangle flips. This helps prevent the production of overly thin or excessively large triangles. Specifically, we repeat the following steps to obtain the subsequent level  and  corresponding map, using the current mesh.

\begin{enumerate}[\hspace{10pt}(1)]
    %\item Pick an edge $e^{\star}$ of the current mesh to collapse and generate a new vertex $v^{\star}$ by QEM, as shown in Figure~\ref{fig:initail_collapse}. Thus, we have two different surface patches $P_{3d}$ and $Q_{3d}$ which only contains the one-ring neighbourhood of $e^{\star}$ and $v^{\star}$, respectively. Flattening the two patches onto the same UV domain, we get two new patches, $P_{2d}$ and $Q_{2d}$. More details can be found in~\cite{liu2020neural}.
    \item Pick an edge $e^{\star}$ of the current mesh to collapse and generate a new vertex $v^{\star}$. The approach is similar to the QEM method, but it differs in that it uses not only approximation error $E_{approx}$ but also a distortion error to determine the priority of the edge. The distortion error refers to the symmetric Dirichlet energy~\cite{intersurface2004} $E_{distort}$ of the map between two surface patches $P_{3d}$ and $Q_{3d}$, where $P_{3d}$ and $Q_{3d}$ contain the one-ring neighbourhood of $e^{\star}$ and $v^{\star}$, respectively, as shown in Figure.~\ref{fig:initail_collapse}. We use a weight $w$ to compute a linear combination of two errors: $E = (1-w)*E_{approx} + w*E_{distort}$.
    To construct such a map, we first flatten $P_{3d}$ and $Q_{3d}$ onto the same UV domain, and then overlay the two parameterizations $P_{2d}$ and $Q_{2d}$ to compute a mutual tessellation. Refer to \cite{disk_para2019} for more details on constructing a map between two disk-topology meshes and measuring the distortion of that map.

    \item Now, each vertex $v$ of $M^{n+1}$ contained in $P_{3d}$ is mapped to a triangle of $Q_{2d}$. We use the point location algorithm~\cite{brown1997robust} to find the index $I(v)$ of the triangle and compute the corresponding barycentric coordinates$(\alpha, \beta)$. We compute $v$'s image $\phi(v)$ in $Q_{2d}$ as follows: 
    \begin{equation}
     \phi(v) = \alpha u_{1}+\beta u_{2}+ (1-\alpha-\beta) u_{3},
    \end{equation}
    where $u_{1}, u_{2}, u_{3}$ are vertex locations of the $I(v)$-th triangle of $Q_{2d}$. We define a triplet $\gamma(v) = (I(v), \alpha, \beta)$ to encode $\phi(v)$, as shown in Figure.~\ref{fig:mapping}(b).

    \item Each edge $e$ of $M^{n+1}$ contained in $P_{3d}$ is fully or partially mapped to a line segment $\psi(e)$ (see Figure.~\ref{fig:mapping}(b)) in $Q_{2d}$. 
    For each $\psi(e)$, we find all the edges of $Q_{2d}$ that intersect it. Assuming that $\psi(e)$ going from vertex $\phi(v_1)$ to vertex $\phi(v_2)$ intersects an edge of $Q_{2d}$ going from vertex $u_{1}$ to vertex $u_{2}$, we solve the following $2\times2$ linear system to get the intersection point:
    \begin{equation}
     \lambda_{1}\phi(v_{1}) + (1-\lambda_{1})\phi(v_{2})  = \lambda_{2} u_{1} + (1-\lambda_2) u_{2},
    \end{equation}
    where $\lambda_{1}$, $\lambda_{2}$ are the linear combination coefficients. All the edge intersection tests can be parallelized. 
    
    \item For each edge-to-edge intersection point $s$, we create a quadruplet that stores two indices of incident edges and two linear combination coefficients denoted by $\tau(s) = (J_{1}, J_{2}, \lambda_{1}, \lambda_{2})$, where $\lambda_{1}$ is associated with the $J_{1}$-th edge of $M^{n+1}$, and $\lambda_{2}$ is associated with the $J_{2}$-th edge of $Q_{2d}$. Now we use a list of quadruplets $\{\tau(s_{1}), \cdots, \tau(s_{n})\}$ to encode $\psi(e)$, where $\{s_i\}_{i=1}^{n}$ are the computed intersection points and are recorded in one halfedge directed from $\phi(v_1)$ to $\phi(v_2)$. Since some of these intersection points may appear on the boundary of $Q_{2d}$ in the following iterations, we record all the quadruplet data in each iteration to judge which edges of $M^{n+1}$ are partially mapped to $Q_{2d}$ in the following iterations, see Figure.~\ref{fig:mapping}(b),(c) and (d).
    \item Update the current mesh locally, replacing $P_{3d}$ with $Q_{3d}$. Meanwhile, we use recorded codes ($\{\gamma\}$ and $\{\tau\}$) to relocate the vertices of $M^{n+1}$ on the newly updated mesh and identify the edge-to-edge intersection points on that mesh as well.
\end{enumerate}

\begin{figure}[htb]
  \centering
  \includegraphics[width=0.9\linewidth]{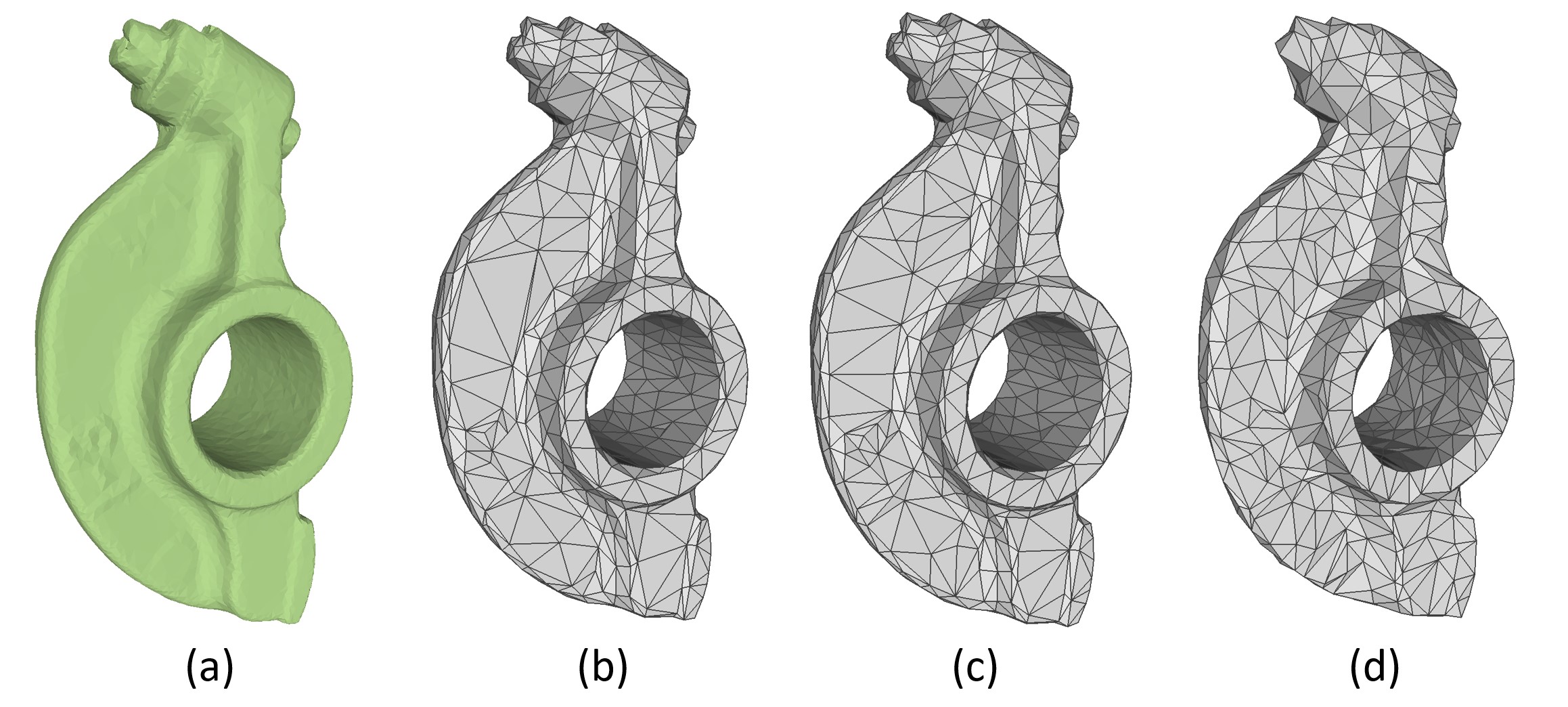}
  \parbox[t]{.8\columnwidth}{\relax }
  \caption{\label{fig:distortion_error}
           For input mesh (a), we visualize the decimated mesh with different values of weight $w$ for combining the approximation error and the distortion error. $w=0$ gives the pure QEM simplification that generates shape-preserving decimated mesh (b), while $w=0.1$ helps the reduction of thin triangles in the decimated result (c), and $w=0.3$ leads to the result with more uniform-areas (d).}
\end{figure}

After the user-specified number of iterations is reached, we arrive at the $n$-th level. At this point, we obtain a meta mesh $M^{n+1,n}$ formed by embedding $M^{n+1}$ on $M^{n}$, which consists of the embedded vertices of $M^{n+1}$, the vertices of $M^{n}$, and the edge-to-edge intersection points. 
\begin{wrapfigure}{r}{0.25\linewidth} 
\includegraphics[width=0.25\columnwidth]{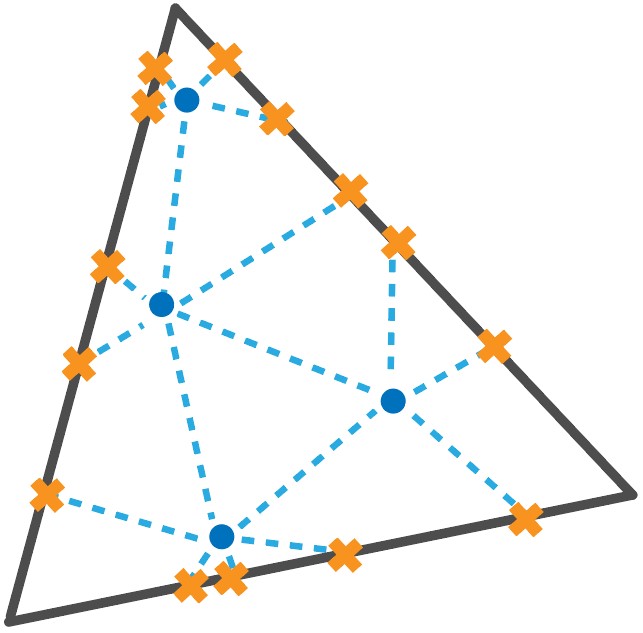}
\end{wrapfigure}
However, we do not yet know the connectivity of $M^{n+1,n}$. To complete $M^{n+1,n}$, we need to extract the overlay polygon between any pair of intersected triangles. Specifically, for each face of $M^{n}$, we use the recorded codes ($\{\gamma\}$ and $\{\tau\}$) to find all the triangles of $M^{n+1}$ that intersect it, and compute the triangle-to-triangle intersection. For more details of the intersection implementation, refer to~\cite{triangle_intersection22}. Moreover, in the step $(2)$ and step $(4)$ of the method described above, we can record additional triplets and quadruplets for embedding $M^{n}$ on $M^{n+1}$. Thus, we obtain another meta mesh $M^{n,n+1}$ formed by embedding $M^{n}$ on $M^{n+1}$ (see Figure.~\ref{fig:mesh_sequence}). Since $M^{n+1,n}$ and $M^{n+1,n}$ have the same connectivity, and each of their faces is a planar convex polygon, the bijective mapping $\mathcal{H}^{n}$ can be naturally defined in a piecewise linear manner.

\textbf{Complexity.} The QEM~\cite{qem97} algorithm takes $\mathcal{O}(\log N)$ time to collapse one edge by removing the least cost from a heap where $N$ is the number of edges; the conformal flattening is a constant cost on top of each edge collapse since valence is bounded, and the edge-to-edge intersections can be implemented in a linear complexity~\cite{triangle_intersection22} on top of each conformal flattening. Thus, one edge collapse actually takes $\mathcal{O}(N)$ time, and the complexity of the entire algorithm is $\mathcal{O}(N^2)$. It is worth pointing out that edge-to-edge intersections can be trivially parallelizable, so our algorithm has the potential to be implemented in the complexity of $\mathcal{O}(N \log N)$.

\section{SPMM-Net}

\begin{figure*}
  \centering
  \includegraphics[width=0.9\textwidth]{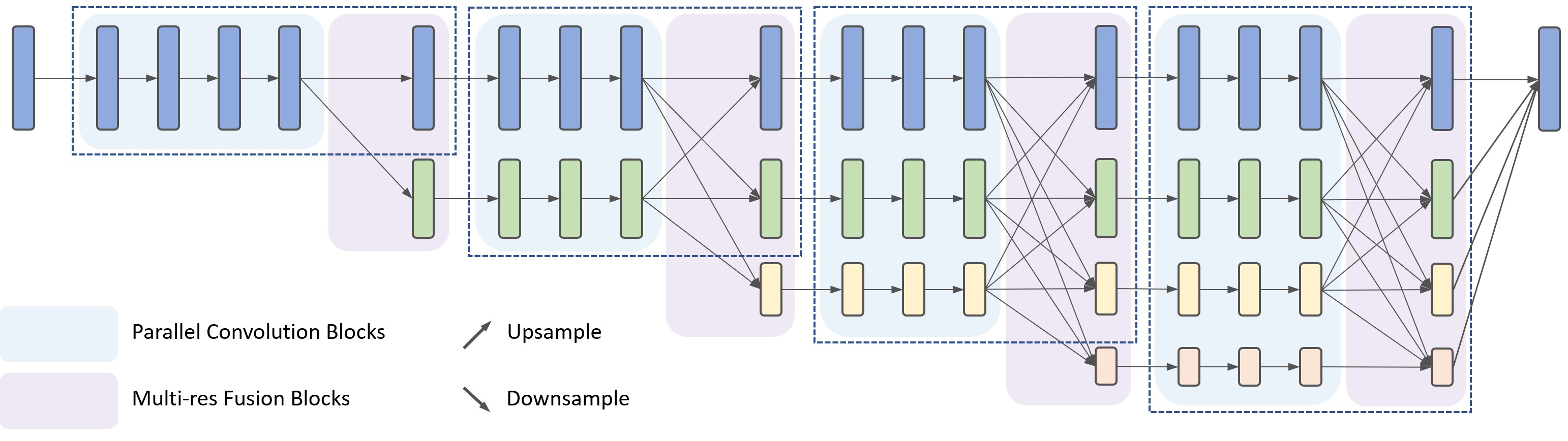}
 \caption{\label{fig:network} The architecture of SPMM-Net is based on HRNet~\cite{HRnet2}. The network maintains the input high-resolution mesh across its stages to obtain spatially precise dense prediction results. The network consists of 4 stages (indicated by the dashed line boxes), each of which comprises multiple parallel convolution blocks and multi-resolution fusion blocks. In the parallel convolution block, features are propagated independently along each resolution level. In the multi-resolution fusion block, the high-resolution data are downsampled, and the low-resolution data are upsampled. The different resolution representations are fused by summation to enable information exchange between different resolution levels.}
\end{figure*}

We are now ready to present our network. It takes high-resolution mesh data as input and aims to learn high-level representations for each face, which can benefit downstream dense prediction tasks. Specially, we utilize the mesh pyramid with bijective surface-to-surface maps to design both high-to-low and low-to-high face feature aggregation schemes that provide similar functionality to pooling and unpooling operations in 2D CNNs. Moreover, we define an order-invariant face-based convolution aggregating neighbor information through layers. With flexible downsampling, upsampling, and convolution operations, we adapt powerful CNN dense prediction architectures~\cite{HRnet1, HRnet2} to irregular 3D surface meshes, thus enabling dense prediction tasks on meshes using SPMM-Net (see Figure.~\ref{fig:network}). The main technical components of our network are detailed in the following subsections.

\subsection{Input features}
Given an input mesh $M$, we first embed the shape and location information for each face in $F$ as the input feature. The input feature to the network for each triangular face is a 10-dimensional vector derived from geometry. It consists of a 7-dimensional global feature that describes the triangular face center $\{ C_{x}, C_{y}, C_{z}\}$, face area $\{a\}$, face normal $\{ N_{x}, N_{y}, N_{z}\}$, and a 3-dimensional vector $\{ k_{0}, k_{1}, k_{2}\}$ that describes the relations between each face and its three neighboring faces, which is calculated as the dot product between vertex normals and the face normal~\cite{Hertz2020deep}. As three vertices of the face are unordered, we remove the ambiguity of the curvature vector by applying the invariant computation:
\begin{equation}
    \nonumber
    \begin{cases}
        g_{0} = k_{0} + k_{1} + k_{2}, \\
        g_{1} = |k_{0} - k_{1}| + |k_{1} - k_{2}|+ |k_{2} - k_{0}|, \\
        g_{2} = |k_{0} - 2k_{1} + k_{2}| + |k_{1} - 2k_{2} + k_{0}|+ |k_{2} - 2k_{0}+k_{1}|
    \end{cases}
\end{equation}
The input 10-dimensional vector $h_{i}^{F}$ that describes the shape and position of the $i$-th face is defined as follows:
\begin{equation}
    h_{i}^{F} = [ C_{x},C_{y},C_{z},a,N_{x},N_{y},N_{z},g_{0},g_{1},g_{2}].
\end{equation}
For convenience, we ignore the use of the index $i$ on the right side of the above equation.

\subsection{Parallel convolution block}

%In the proposed network, the face features are propagated independently along each resolution level in the parallel convolution block. Within each parallel convolution block, regular convolutions with identical residual learning strategy~\cite{resnet2016} are performed over each level. Each convolution block is defined to consist of 4 residual units, and a residual unit contains 2 face convolution layers, 2 batch normalization layers, and 2 ReLU layers.
\textbf{Convolution operation.}
In a watertight triangle mesh without boundary edges, each face is adjacent to 3 neighbor faces. Motivated by the convolution operation defined in CNN, which takes a fixed number of 9-pixel representations for each calculation, the basic convolution pattern on a triangular mesh for face $f_{i}$ is formed by aggregating the features of its 1-ring face-neighborhood $\mathcal{N}\left(f_{i}\right)$. Unlike images, the neighboring face set $\mathcal{N}\left(f_{i}\right)$ in the mesh is unordered, yet the order variance could cause ambiguity in the calculation. In our network, we follow the setting from SubdivNet~\cite{hu2021subdivision} that utilizes summation to achieve invariance and define the convolution operation on face $f_{i}$ as follows:
\begin{equation}
    h_{i}^{F} = w_{0} h_{i}^{F} + w_{1}\sum_{j=1}^{3}h_{i_j}^{F} + w_{2}\sum_{j=1}^{3}|h_{i_{j+1}}^{F}-h_{i_{j}}^{F}| + w_{3}\sum_{j=1}^{3}|h_{i}^{F}-h_{i_{j}}^{F}|,
\end{equation}
where $\{h_{i_j}^{F}\}_{j=1}^{4}$ are the face features of $\mathcal{N}\left(f_{i}\right)$ in counter-clockwise order, $h_{i_4}^{F}=h_{i_1}^{F}$, and $\{w_{i} | w_{i}\in \mathbb{R}^{L \times D}, i=0,1,2,3\}$ are trainable parameters with $L$ and $D$ being input and output channels, respectively. 

\begin{wrapfigure}{r}{0.25\linewidth}
    \includegraphics[width=0.25\columnwidth]{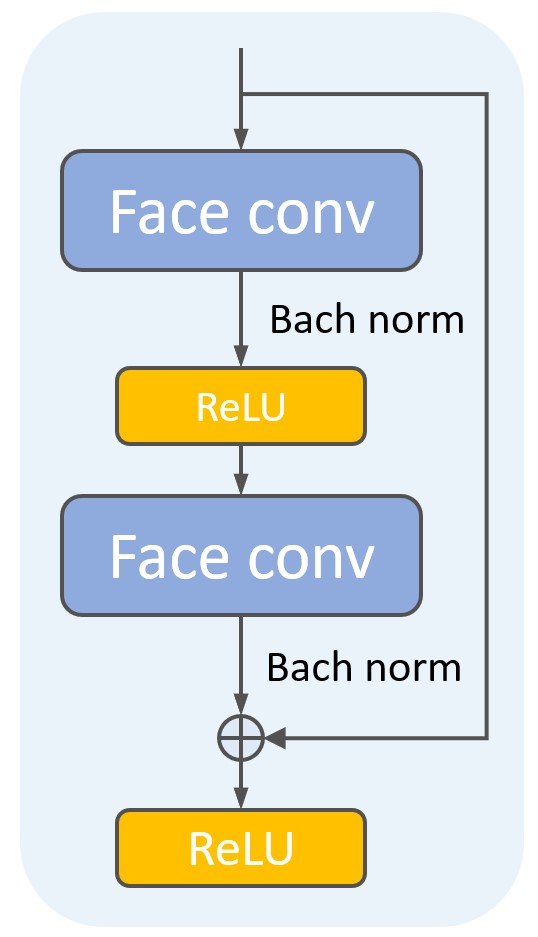}
  \label{fig:fusion}
\end{wrapfigure}

In the proposed network, the face features are propagated independently along each resolution level through parallel convolution blocks (see light blue blocks in Figure.~\ref{fig:network}). Within each parallel convolution block, regular convolutions with identical residual learning strategy~\cite{resnet2016} are performed over each level. Each convolution block is defined to consist of four residual units, and a residual unit contains two face convolution layers, two batch normalization layers, and two ReLU layers.

\subsection{Multi-resolution fusion block}

Multi-resolution fusion exchanges information between different resolution levels, which is frequently used in learning architectures for 2D dense prediction. The low-resolution features are likely to contain global semantic information, and high-resolution features are sensitive to local geometry and precise spatial positions. By combining these information, the fused features would be semantically richer and spatially more precise, which thus benefits the downstream tasks. 

\begin{wrapfigure}{r}{0.3\linewidth}
    \includegraphics[width=0.3\columnwidth]{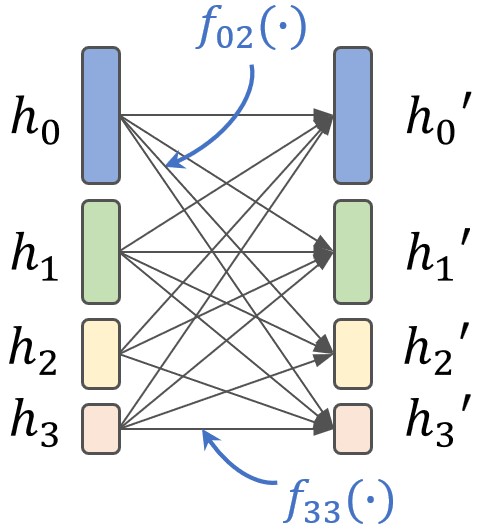}
\end{wrapfigure}

In SPMM-Net, the multi-resolution feature fusion (see lilac blocks in Figure.~\ref{fig:network}) is performed after each multi-resolution convolution block. The figure on the right illustrates a four-level fusion block, where the input features to the fusion block are: $\left\{h_{r}\right\}_{r=0}^{3}$ with $r$ being the index of resolution level, and outputs of the fusion block $\left\{h_{r}^{'}\right\}_{r=0}^{3}$ are computed as the average of transformed features: $h_{r}^{'} = \frac{1}{4}\cdot \left( T_{0 r}\left( h_{0} \right) + T_{1 r}\left( h_{1} \right) + T_{2 r}\left( h_{2} \right) + T_{3 r}\left( h_{3} \right) \right)$. The transform function $T_{x r}(\cdot)$ is decided by input resolution level index $x$ and output resolution level index $r$. If $x < r$, the input features are downsampled for $r-x$ rounds; If $x > r$, the input features are upsampled for $x-r$ rounds; While $T_{xr}(h) = h$ when $x = r$.

Next, we will demonstrate the downsampling and upsampling opertations between two neighboring levels $M^{n+1}$ and $M^{n}$. With the constructed bijective map $\mathcal{H}_n: M^{n+1} \rightarrow M^{n}$ in the previous section, we can easily adapt the pooling and unpooling operations in images to irregular mesh data and then achieve feature fusion. Our downsampling and upsampling operations utilize mesh self-parameterization as clustering tools and precisely aggregate the representation through the layer.

%The constructed piece-wise linear bijective surface mappings are used to achieve multi-resolution feature fusion for learning on meshes in our network. In particular, with the defined sequential bijective mappings, we can easily adapt the pooling and unpooling operations in images to irregular mesh data, and then achieve feature fusion. Our downsampling and upsampling operations utilize the mesh self-parameterization as clustering information and precisely aggregate the representation through the layer. Compared with the operations defined in SubdivNet\cite{hu2021subdivision}, which only consideres the shape connectivity (four face to one) during pooling and unpooling, our method is more general as we take the face area into consideration.

\begin{figure}[htb]
  \centering
  \includegraphics[width=0.9\linewidth]{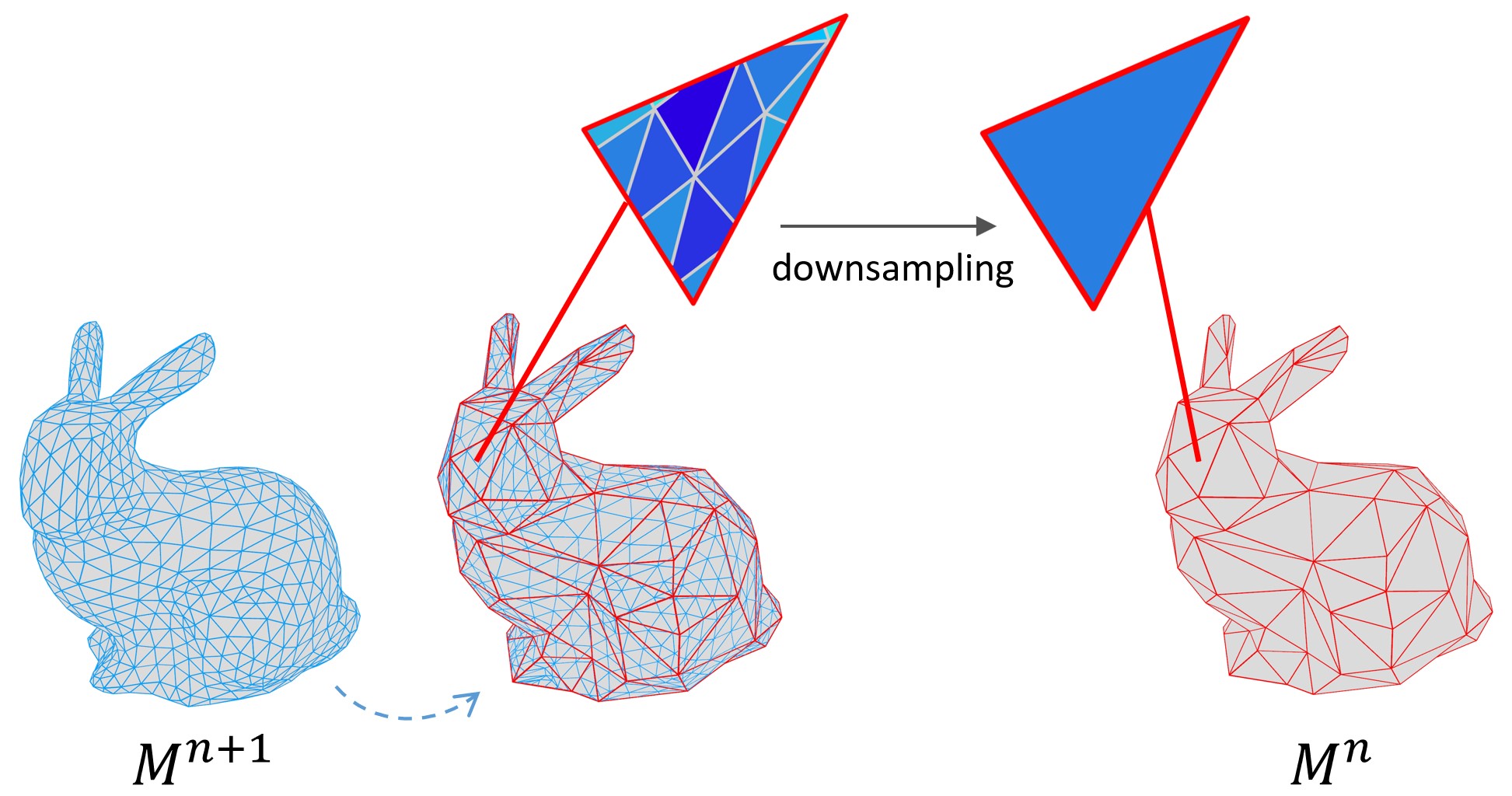}
  \parbox[t]{.9\columnwidth}{\relax }
  \caption{\label{fig:downsampling}
           Visualization of the downsampling process: A cluster of faces from mesh $M^{n+1}$ contributes to one face in the decimated mesh $M^{n}$, where the contribution score $\alpha$ is determined by the intersection area. }
\end{figure}

\textbf{Downsampling.}
Firstly, we compute a sparse face mapping matrix $A^{n}=\{a^n_{i,j}\}\in \mathbb{R}^{|F^{n+1}|\times|F^n|}$, where $a^n_{i,j}$ measures the overlap ratio between the face $f^{n}_{j}$ of $M^{n}$ and the image $\mathcal{H}^{n}(f_{i}^{n+1})$ of the face $f_{i}^{n+1}$ of $M^{n+1}$ (see Figure.~\ref{fig:downsampling}) via the following equation:
\begin{equation}
    a^n_{i,j}=\frac{S\left( \mathcal{H}^{n}(f_{i}^{n+1}) \bigcap {f^n_{j}} \right)}{S\left( f^n_j \right)}.
\end{equation}
$S(\cdot)$ computes the planar polygon area. Then we propose a weighted average fusion method to define the downsampling operation from mesh $M^{n+1}$ to mesh $M^{n}$ as follows: 
\begin{equation}
h_{j,n}^{F} = \frac{\sum_i a^n_{i,j} \cdot h_{i,n+1}^{F}}{\sum_i a^n_{i,j}},
\end{equation}
where $h_{i,n+1}^{F}$ represents the feature vector for face $f_{i}^{n+1}$ and $h_{j,n}^{F}$ represents the pooled feature vector for face $f_{j}^{n}$.

%With obtained bijective map $\mathcal{H}_n: M^{n+1} \rightarrow M^{n}$, we project mesh $M^{n+1}$ on mesh $M^{n}$. A sparse face mapping matrix $\mathbf{A^n}=\{a^n_{i,j}\}\in \mathcal{R}^{|F^{n+1}|\times|F^n|}$ is obtained, where $a^n_{i,j}$ measures the overlap ratio of area of face $g^n_i=(\mathcal{H}^{n})^{-1}(f^{n+1}_{i})$ in $M^{n+1}$ that intersects face $f^n_{j}$ in $M_{n}$ on the projected plane (see Figure.\ref{fig:calculate_intersections}).

In practice, we obtain the multi-resolution meshes by sequentially applying a fixed number of edge collapse till the number of vertices reaches the target for each resolution level. As mesh $M^{n+1}$ with $|F^{n+1}|$ faces has been decimated to $M^n$ with $|F^{n}|$ faces, the downsampling rate between two neighboring levels is defined as $|F^{n+1}|/|F^n|$. The downsampling rate is a hyper-parameter for our network; we decide it together by the number of faces of the finest mesh and base mesh, and the height of the pyramid. In our experiments, it is empirically set to $4$ for training datasets. 

\textbf{Upsampling.} The upsampling operation can be seen as the reverse of downsampling. With the sparse face mapping matrix $A^{n}$, we define the area-aware upsampling operation as follows:
\begin{equation}
h_{i,n+1}^{F} = \frac{\sum_j a^n_{i,j} \cdot h_{j,n}^{F}}{\sum_j a^n_{i,j}}.
\end{equation}
%The upsampling operation can be defined as the reverse of downsampling. Projecting mesh $M^n$ on mesh $M^{n+1}$, we derive the sparse face mapping $U^{n+1} = \{u^n_{i,j}\}\in \mathcal{R}^{|F^{n}|\times|F^{n+1}|}$ that measures the ratio of area that face $f^{n}_i$ intersects face $\mathcal{H}^{n}(f^{n+1}_j)$ on mesh $M^{n+1}$. The area-aware upsampling operation is defined as follows:
%\begin{equation}
%h_{f}^{i} = \frac{\sum_i u^{n+1}_{i,j} \cdot h_{f}^{j}}{\sum_i %u^{n+1}_{i,j}}.
%\end{equation}

\begin{figure*}[htbp!]
  \centering
  \includegraphics[width=0.86\linewidth]{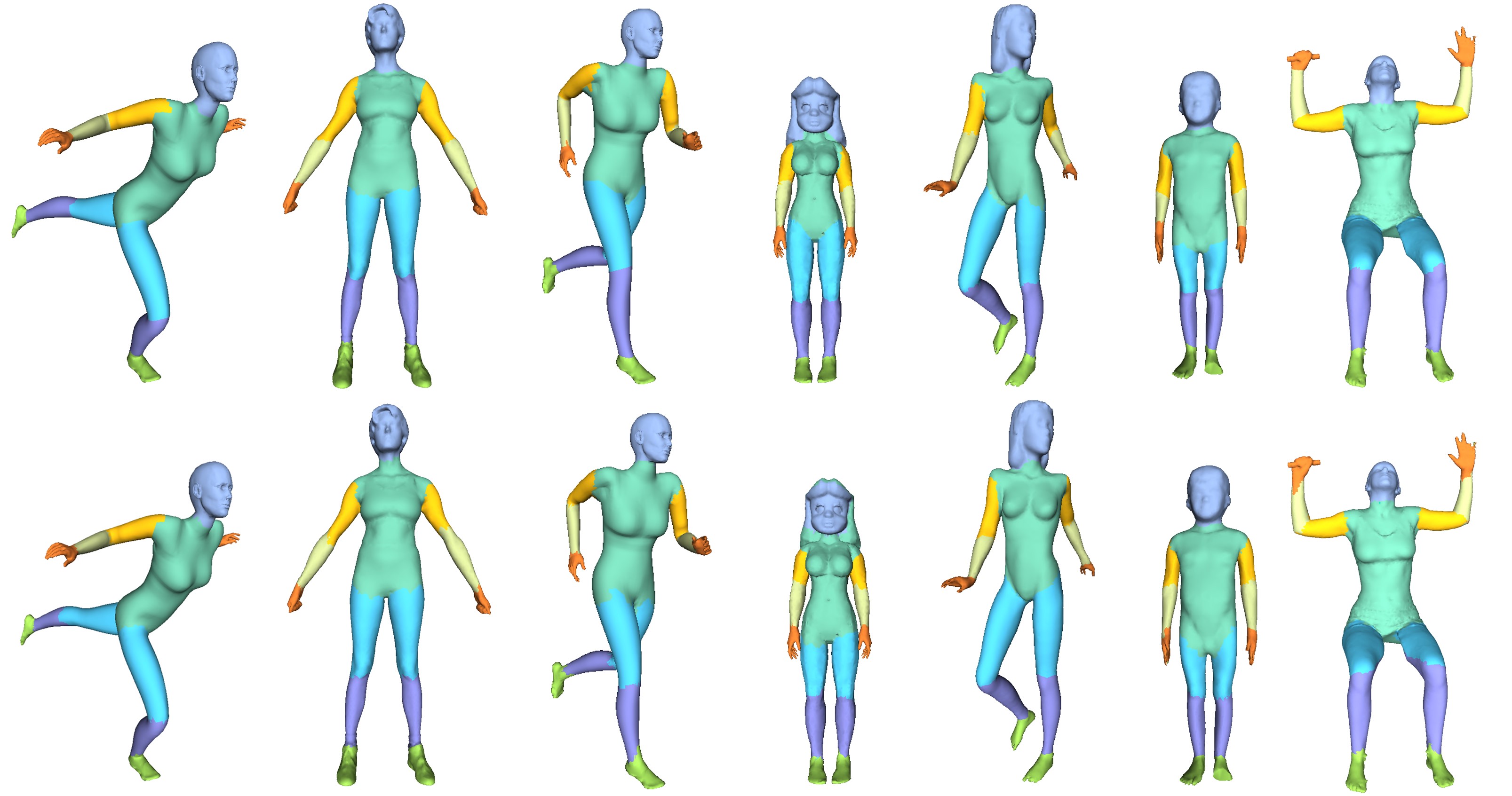}
  \caption{\label{fig:human_seg}Visualization of segmentation results. Top: SPMM-Net segmentation results. Bottom: ground truth.}
\end{figure*}

From the above equation, we can see that if face image $\mathcal{H}(f_{i}^{n+1})$ is fully contained in face $f_{j}^{n}$, then feature $h_{j,n}^{F}$ of face $f_{j}^{n}$ is directly assigned to face $f_{i}^{n+1}$. As the sampling rate increases, more faces of mesh $M^{n+1}$ end up sharing the same upsampled feature, resulting in the process resembling nearest neighbor sampling. Thus, the generated feature maps could be very sparse and unlikely to depict semantically consistent regions~\cite{Lu_2019_ICCV}. To improve this, we propose a more effective upsample operation based on barycentric interpolation. We first compute a feature $h_{i,n}^{V}$ for each vertex $v_{i}^{n}$ of $M^{n}$ by averaging its incident faces features $h_{j,n}^{F}$:
\begin{equation}
h_{i,n}^{V} = \frac{1}{\mathcal{N}(v_i^n)} \sum_{j \in \mathcal{N}(v_i^n)}h_{j,n}^{F},
\end{equation}
where $\mathcal{N}(v_i^n)$ is the 1-ring vertex-neighbourhood of $v_{i}^{n}$. Then, we interpolate the vertex feature for each vertex $v_{i}^{n+1}$ of $M^{n+1}$ via:
\begin{equation}
h_{i,n+1}^{V} = \alpha h_{i_1,n}^{V} + \beta h_{i_2,n}^{V} + (1-\alpha-\beta) h_{i_3,n}^{V}.
\end{equation}
$\{h_{i_1}^{V},h_{i_2}^{V},h_{i_3}^{V}\}$ are vertex features of the triangle where $\mathcal{H}^{n}(v_{i}^{n+1})$ lies and $(\alpha,\beta)$ are the corresponding barycentric coordinates. Lastly, we compute each face feature $h_{i,n+1}^{F}$ of $M^{n+1}$ as the average of its incident vertex features:
\begin{equation}
h_{i,n+1}^{F} = \frac{1}{3} \sum_{j=1}^{3}h_{i_j,n+1}^{V}.
\end{equation}

%However, in practice, we have observed that the above approach can lead to homogeneous feature propagation when the down/up sampling rate is high. This is because when the face $f_{j}^{n}$ on mesh $M^{n}$ is completely overlapped by a face $f_{i}^{n+1}$ on mesh $M^{n+1}$, the feature of face $f_{i}^{n+1}$ is directly assigned to face $f_{j}^{n}$. As the sampling rate increases, more faces on mesh $M^{n}$ end up sharing the same upsampled feature, which can result in the process resembling nearest neighbor sampling. To avoid this, we propose a barycentric interpolation upsample process that untilizes the vertex-to-surface map ($M^{n+1} \rightarrow M^{n}$). Firstly, we compute the representation of each vertex $h_{v}^{i}$ on mesh $M^{n+1}$ by averaging the features $h_{f}^{j}$ of its incident faces. Then, using the vertex-to-surface map, we interpolate the vertex feature $h_{v}^{i}$ on the image of mesh $M^{n}$ as $h_{v}^{i}=\sum_{k=1}^{3}\lambda_{k}h_{v}^{k}$, where $(\lambda_{1},\lambda_{2},\lambda_{3})$ denote the barycentric coordinates and $h_{v}^{k}$ are the corresponding vertex features of the related triangle. Lastly, the feature $h_{f}^{i}$ of face $f_{i}^{n+1}$ on mesh $M^{n+1}$ is calculated as the average of adjacent vertices' features.

%%%%%%%%%%%%%%%%%%%%%%%%%%%%%%%%%%%%%%%%%%
%%%%%%%%%%%%%%%%%%%%%%%%%%%%%%%%%%%%%%%%%%
\section{Experiments}
This section reports our experimental results and highlights the effectiveness of our proposed framework. We conducted evaluations on dense prediction for meshes, including mesh segmentation and shape correspondence. We also performed comparisons of our approach with the state-of-the-art and assessed the key components of our network. The results provide convincing evidence of the superior performance of our framework in accomplishing these tasks.

\subsection{Implementation}
\textbf{Dataset.}
Since our method targeted dense prediction tasks, we required the input mesh to contain more than 10,000 faces. Additionally, the input mesh should be watertight, so we utilized hole-filling  algorithms to repair meshes that contain holes or boundaries. Subsequently, we constructed a set of multi-resolution meshes and a collection of bijective surface-to-surface maps using the method described in Section 3. The weight $w$ for linear combination the approximation error $E_{approx}$ and the distortion error $E_{distort}$ is set to 0.1. Our experiments encompassed several datasets, including the human body segmentation dataset~\cite{Maron2017}, COSEG dataset~\cite{coseg2012}, and FAUST shape correspondence dataset~\cite{faust2014}. 
The meshes in the COSEG dataset are of low-resolution, and therefore, we processed meshes in the dataset and generated high-resolution data.

%Since our method requires the input mesh to be watertight, we first use filling holes algorithms to repair meshes containing holes or boundaries. Subsequently, we constructed a set of multi-resolution meshes and a collection of bijective surface-to-surface maps using the method described in Section 3. For datasets with meshes that vary significantly in face numbers, particularly those with low-resolution inputs, we remeshed the models to ensure that the meshes in the same dataset have roughly the same number of elements. Additionally, to demonstrate the effectiveness of our network in dealing with dense prediction problems, we ensured that the input mesh should contain more than 10,000 faces. 

%\textbf{Feature propagation.} 
%To derive the prediction for the original low-resolution mesh $M^{O}$, we incorporated a feature transfer layer to propagate face representations from the high-resolution input $M^{K}$ to $M^{O}$ before making final predictions. Specifically, we first computed the vertex representation for $v_{i}^{K}$ on the high-resolution shape by averaging the features of its incident faces. Next, we projected the vertices $v_{j}^{O}$ of the original mesh onto the high-resolution shape, which enabled us to obtain the face index and barycentric coordinates of the projection. We then proceeded to interpolate the feature of $v_{j}^{O}$ and compute the representation of the triangle face $f_{j}^{O}$ by averaging the features of its three vertices. These steps enabled us to transfer the features from high-resolution mesh to low-resolution data.

\textbf{Augmentation.} Given a high-resolution input $M^{K}$, we generated a few variations of decimated mesh pyramids to reduce the variance introduced by using different edge collapse orderings. That is, we constructed several collections of surface-to-surface maps for each input mesh. The procedure helps to reduce the influence of the decimation and improve the robustness of the model. However, as the augmentation also increases the training time and computational cost, it is necessary to balance the benefits against these considerations.

Following the process outlined in~\cite{MeshCNN2019}, we scaled each input mesh to fit into a unit cube and applied a random scaling with a normal distribution $\mu = 1$ and $\sigma = 0.1$, in order to reduce the network's sensitivity to the size of the model. As some datasets contain meshes with different orientations, we further applied a random rotation along the $[x, y, z]$ axes with Euler angles of $[0, \pi/2, \pi, 3\pi/2]$, in accordance with the settings in~\cite{hu2021subdivision}. These measures aimed to increase the diversity of the training data and enhance the robustness and generalization performance of the proposed model.

\textbf{Networks.} 
The proposed framework employs HRNetV2p~\cite{HRnet2} as the backbone feature extractor. In our framework, a multi-resolution block consists of a parallel convolution block and a multi-resolution fusion block. Each parallel convolution block comprises four residual units, and each residual unit is composed of two face convolution layers, two batch normalization layers, and two ReLU layers. The implemented network contains four resolution levels and four stages, with the number of feature channels for each resolution level set to $[C, 2C, 4C, 8C]$. The network starts with four bottleneck units in the first stage, followed by a downsampling operation. Consistent with the settings in HRNetV2p, the second, third, and fourth stages of the network contain one, four, and three multi-resolution blocks, respectively. The outputs from the multi-resolution levels are concatenated to generate a $15C$-dimensional vector, which is passed through a linear classifier or regressor to make predictions. This design leverages the strengths of the HRNet architecture and enables effective feature extraction across different resolution levels.

\subsection{Segmentation tasks}

We utilized the HRNetV2p-W32 architecture, with $32$ indicating the number of channels of the high-resolution convolutions, and 64, 128, 256 being the number of channels for the 2nd, 3rd, and 4th resolution levels, respectively. To facilitate cross-resolution feature fusion, we employed area-weighed downsampling and barycentric upsampling techniques. For the segmentation tasks, we used the cross-entropy loss function. The SGD optimizer was used with an initial learning rate of $2e^{-2}$, momentum of 0.9, and weight decay of 0.0005. The network was trained for 200 epochs with a batch size of 4 on NVIDIA Quadro RTX 5000, for $14$ hours. Our network takes high-resolution meshes containing approximately 16,384 faces per mesh as input. We constructed the mesh pyramid by sequentially reducing the number of faces to 4,096, 1,024, and 256, with a downsampling rate of 4, resulting in a mesh pyramid with a depth of 4.

\textbf{Human body segmentation dataset.} We evaluated our SPMM-Net on the human body segmentation dataset~\cite{Maron2017}, which aims to predict the probability of each triangular face belonging to a specific part of the mesh. The dataset consists of 381 training human models and 18 testing models from various datasets, with each model segmented into 8 labels. Different from SubdivNet~\cite{hu2021subdivision}, we didn't generate multiple instances of each input mesh, as our method directly deals with irregular input data and provides more stable and robust results.

To ensure a fair comparison with other methods, we presented the segmentation results on the original resolution human body segmentation dataset. From Table.~\ref{table:human_segmentation}, we achieved comparable segmentation results with other state-of-the-art methods. Some segmentation results are visualized in Figure.~\ref{fig:human_seg}. It's worth noting that the highest accuracy reported in SubdivNet~\cite{hu2021subdivision} used the majority voting strategy, which involves remeshing the test data 10 times and using the predictions from different remeshed shapes to make the final prediction. However, the strategy increases the training time and computation cost by $\times 10$ time.

\begin{table}[th]
\caption{Segmentation accuracy on human body segmentation dataset~\cite{Maron2017}.}
\vspace*{3mm}
  \centering
  \label{table:human_segmentation}
  \begin{tabular}{ r@{}l  r }
    \hline\hline %inserts double horizontal lines
    \multicolumn{2}{c}{\textbf{Method}} & 
    \multicolumn{1}{c}{\textbf{Accuracy}} \\
    \hline % inserts single horizontal line
    & PointNet++~\cite{pointnet++}          & $82.3\%$~~~                   \\
    & PD-MeshNet~\cite{PDMeshNet2020}       & $86.9\%$~~~                   \\
    & DiffusionNet~\cite{DiffusionNet2022}  & $91.7\%$~~~                   \\
    & SubdivNet~\cite{hu2021subdivision}    & $91.1\%$~~~                   \\
    & SPMM-Net & $\textbf{91.9}\%$~~~                                       \\
    \hline % inserts single horizontal line
  \end{tabular}
\end{table}

\begin{figure}[htb]
  \centering
  \includegraphics[width=0.9\linewidth]{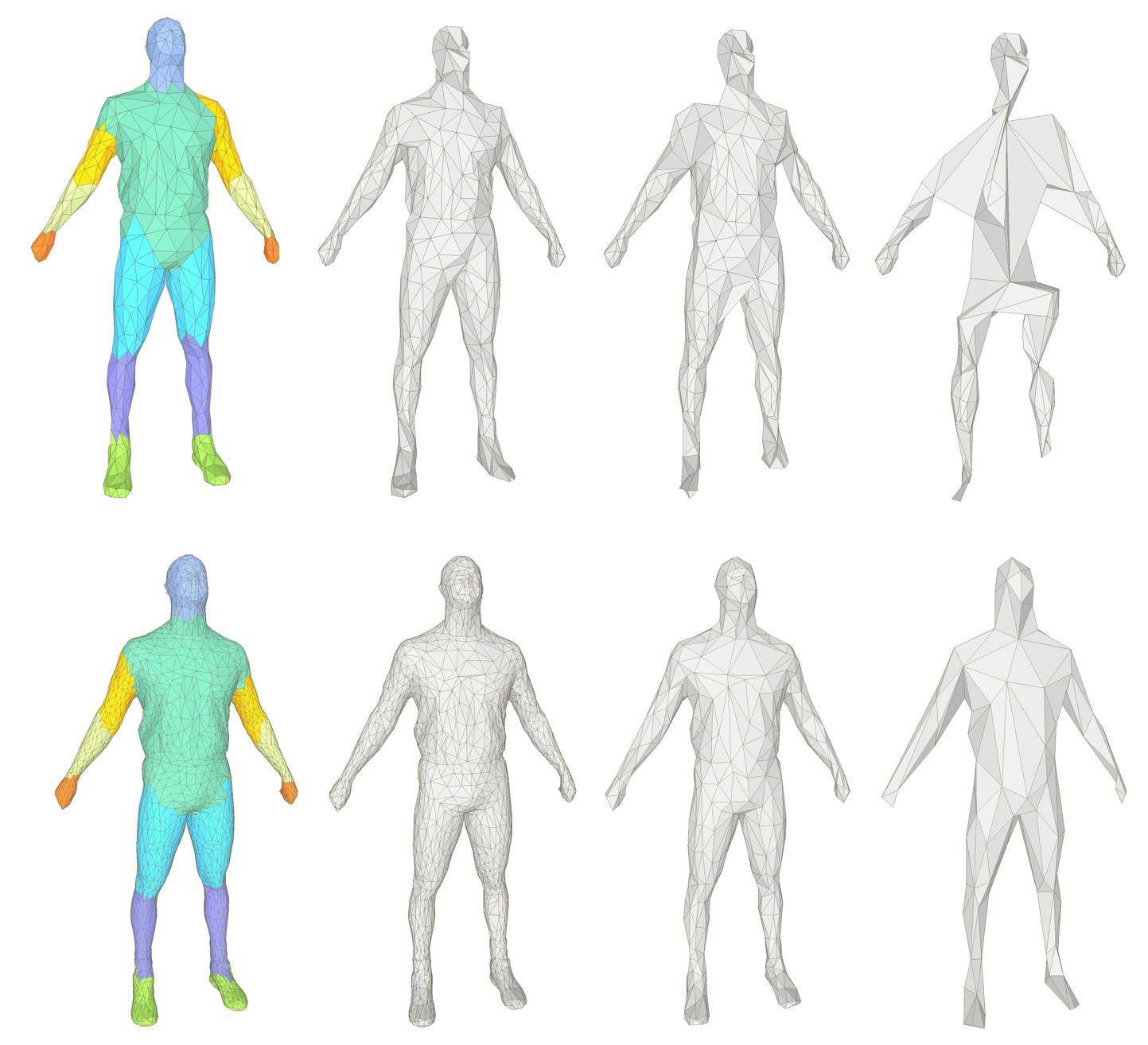}
  \caption{\label{fig:human_comparision}  Segmentation results on the Human body dataset using MeshCNN~\cite{MeshCNN2019} (Top left) and SPMMNet (Bottom left). The results of intermediate pooling operations are shown in sequence. We observe smaller distortions of the pooled meshes with our method.}
\end{figure}

Figure~\ref{fig:human_comparision} displays a qualitative comparison between the segmentation results obtained by MeshCNN~\cite{MeshCNN2019} and SPMMNet. Despite taking input meshes of different resolutions, both methods performed three pooling operations on the input meshes to obtain a base mesh consisting of approximately 300 to 400 faces. We observe that our method better preserves the human shape and results in higher segmentation accuracy. Notably, the shoulders of the model underwent significant changes in the base mesh generated by MeshCNN's pooling method, leading to a drop in segmentation accuracy in this region.

\begin{figure}[htb]
  \centering
  \includegraphics[width=0.95\linewidth]{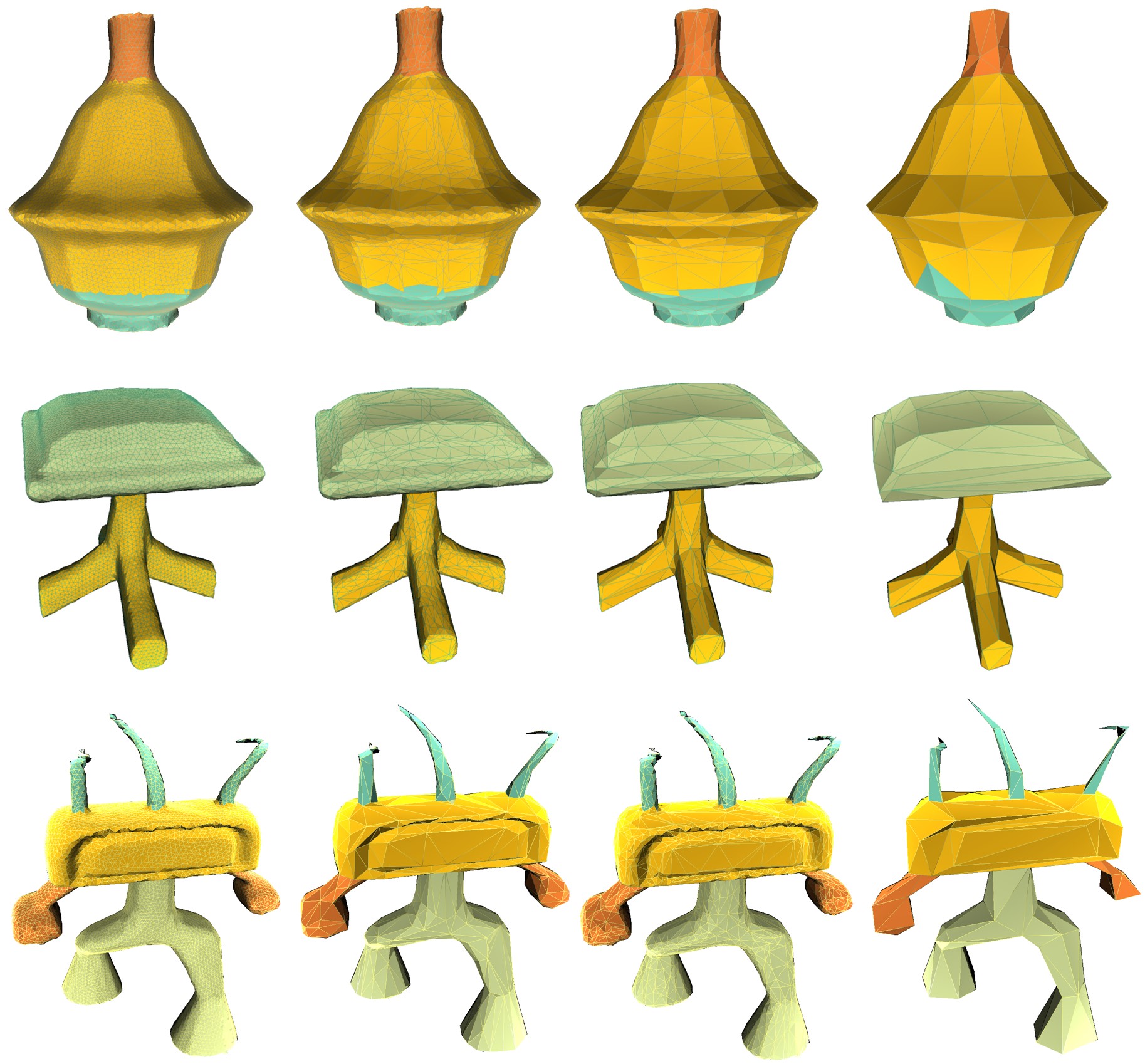}
  \caption{\label{fig:coseg}
           Segmentation results on COSEG dataset~\cite{coseg2012}. The segmentation prediction of high-resolution input is shown on the left, followed by the decimated meshes after each downsampling operation. For visualization purpose, we propagated the prediction on high-resolution mesh to intermediate levels. }
\end{figure}

\textbf{COSEG dataset.} 
We also evaluated our SPMM-Net on the COSEG dataset~\cite{coseg2012}. The dataset contains three large sets of meshes, including 200 alien models, 300 vase models, and 400 chair models. As original COSEG datasets contain non-manifold meshes, we fixed the meshes and generated corresponding high-resolution data. The training was conducted on three datasets independently. We used the soft boundary labels~\cite{PDMeshNet2020} to measure the prediction of each face that is incident to the boundary. Quantitative results are shown in Table.~\ref{table:coseg}. It can be seen that our method performs the best in all categories. We believe that this is due to the uniform propagation scheme through multi-resolution levels. To further show this, we visualize the decimated mesh sequences of the test data in Figure.~\ref{fig:coseg}.

\begin{table}[ht]
\centering % used for centering table
\caption{Segmentation accuracy on COSEG dataset~\cite{coseg2012}.} % title of Table
\vspace*{3mm}
\begin{tabular}{c c c c} % centered columns (4 columns)
\hline\hline %inserts double horizontal lines
Method & Vase & Chair & Alien \\ [0.5ex] % inserts table
%heading
\hline % inserts single horizontal line
MeshCNN~\cite{MeshCNN2019} & $85.2\%$~~~ & $92.8\%$~~~ & $94.4\%$~~~        \\
PD-MeshNet~\cite{PDMeshNet2020} & $81.6\%$~~~ & $90.0\%$~~~ & $89.0\%$~~~   \\
SubdivNet~\cite{hu2021subdivision} & $96.7\%$~~~ & $96.7\%$~~~ & $97.3\%$~~~\\
SPMM-Net & $\textbf{97.0}\%$~~~ & $\textbf{97.2}\%$~~~ & $\textbf{97.6}\%$~~~   \\
\hline %inserts single line
\end{tabular}
\label{table:coseg} % is used to refer this table in the text
\end{table}

\begin{figure*}
  \centering
  \includegraphics[width=0.9
  \textwidth]{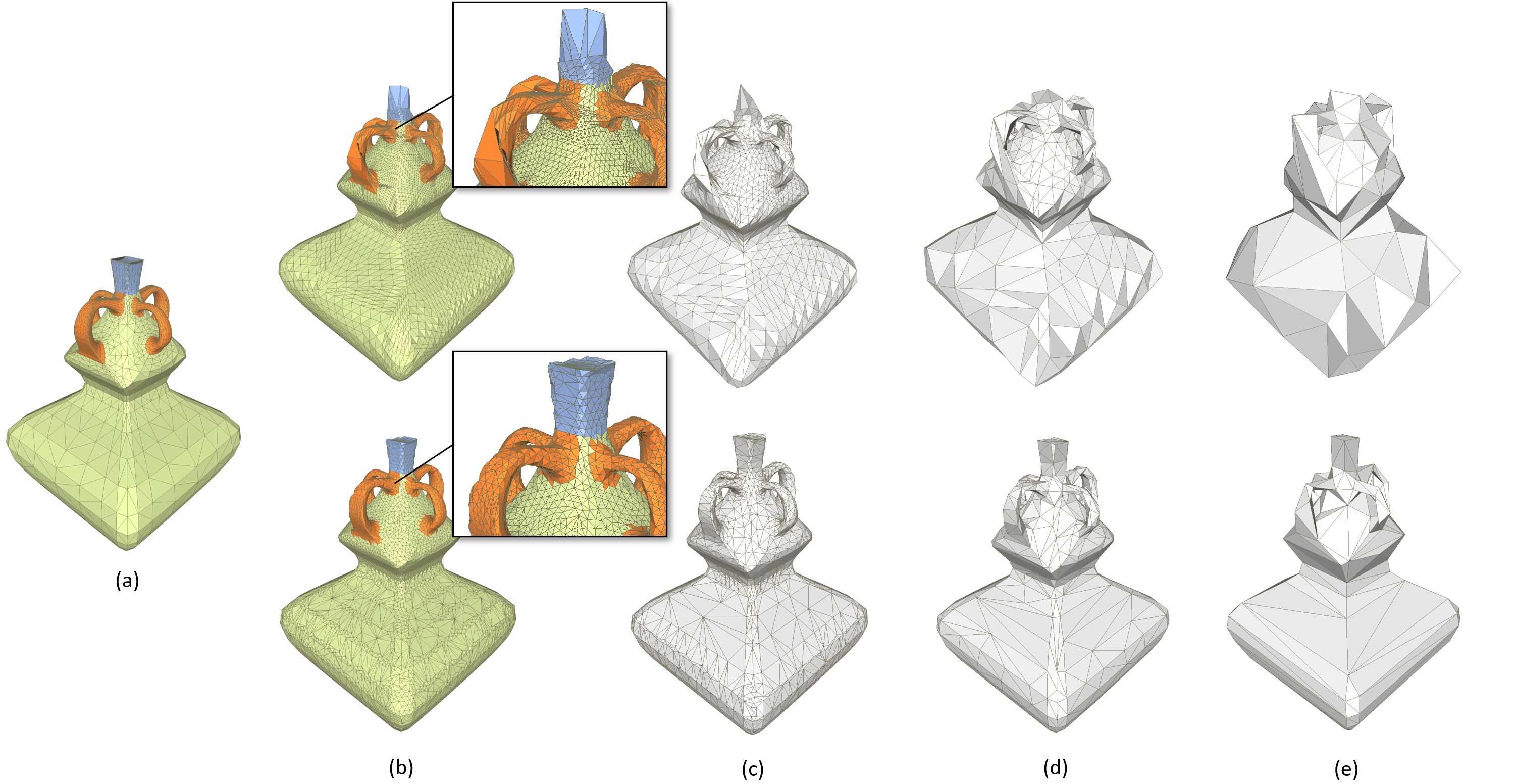}
 \caption{\label{fig:decimation} 
        The mesh pyramids constructed by SubdivNet~\cite{hu2021subdivision} (first row) and SPMM-Net (bottom row). (a) Low-resolution ground truth. (b) High-resolution inputs to the network with 16384 faces. (c)-(e) Decimated meshes contain 4096, 1024, and 256 faces, respectively. SubdivNet decimates the original mesh to base size (256 faces) first and then subdivides the base mesh to create loop connectivity, while our method remeshes the input first and decimates the high-resolution mesh level by level to better preserve the original shape.}
\end{figure*}

Moreover, due to the loop connectivity requirement in SubdivNet, creating a mesh pyramid that precisely describes the original shape is a demanding task. However, as shown in Figure.~\ref{fig:decimation}, our method is flexible to connectivity. Therefore, we can generate a mesh pyramid with higher quality that enhances the learning process.

\begin{figure*}[ht]
  \centering
  \includegraphics[width=0.9\textwidth]{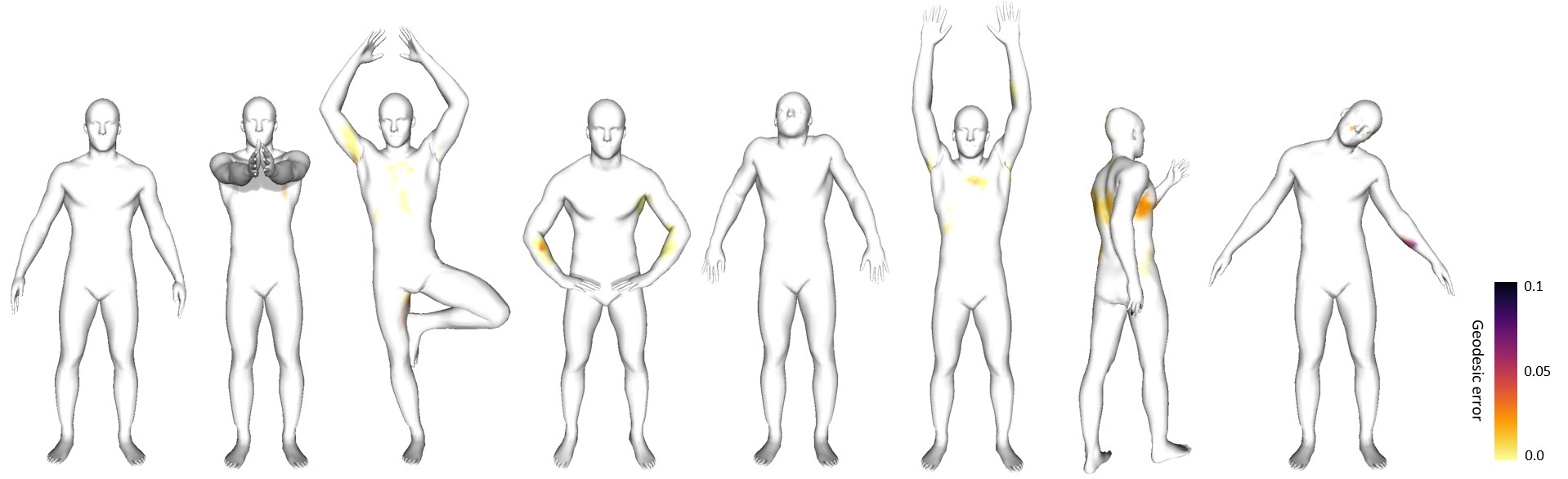}
 \caption{\label{fig:correspondence} 
        Shape correspondence for FAUST dataset~\cite{faust2014}.}
\end{figure*}

\subsection{Shape correspondence tasks}
%\textbf{FAUST dataset} 
We validated our method on Faust~\cite{faust2014} shape correspondence and compared the results with approaches to~\cite{MoNet2017}\cite{SplineCNN2018}\cite{ChebNet} that employ the Princeton benchmark protocol\cite{vladimir2011}. The shape correspondence task in these works refers to the task of labeling each vertex of the input shape to the corresponding vertex of the reference shape. The Faust dataset contains 100 processed scanned human body shapes with ten different poses. Each shape has 6890 nodes and 13776 faces. The first 80 shapes from the dataset are used for training and the rest 20 shapes for testing. The ground truth of the dataset is given implicitly, as each shape has the same connectivity. We took the original meshes as input and constructed the mesh pyramids with number of faces equals to 3444, 861, and 216, with downsampling rate equals to 4.

As our SPMM-Net deals with the face element, we formulated the correspondence problem as a face labeling task, where each face was assigned a 13776-dimensional vector as ground truth.
We employed the HRNetV2p-W48 architecture for this pretty dense task, meaning the resolution level 1st, 2nd, 3rd, and 4th contains 48, 96, 192, 384 convolution channels, respetively. We used the area-weighed downsampling and barycentric upsampling for feature fusion, and employed a soft error loss~\cite{Litany2017DeepFM}. For each input mesh, we first calculated the geodesic distance between the centers of any pair of triangle faces, then, for predictions that are geodesically far away from the correct face, we punished them stronger. We used the same SGD optimizer with segmentation tasks. The SGD optimizer was used with an initial learning rate of $2e^{-2}$, momentum of 0.9, and weight decay of 0.0005. The network was trained for 200 epochs with a batch size of 4 on NVIDIA Quadro RTX 5000, for $8$ hours.

The correspondence error is calculated as the normalized geodesic distance between the predicted face center and the ground truth face center. To measure the correspondence quality, the Princeton benchmark protocol counts the percentage of derived correspondence within a geodesic radius $r$ around the correct face center. Our method achieves state-of-the-art results compared with the methods using the same protocol (See Table.\ref{table:correspondence}). The derived errors are visualized in Figure.~\ref{fig:correspondence}, and we can see that the second to the right body presents a higher geodesic error as the posture is rarely seen in the dataset.

\begin{table}[ht]
\centering % used for centering table
\caption{Correspondence accuracy on FAUST dataset\cite{faust2014}.} % title of Table
\vspace*{3mm}
\begin{tabular}{c c c c} % centered columns (4 columns)
\hline\hline %inserts double horizontal lines
Method & Accuracy (r=0) & Accuracy (r=0.01) \\ [0.5ex] % inserts table
%heading
\hline % inserts single horizontal line
MoNet~\cite{MoNet2017}          & $88.20\%$~~~ & $92.35\%$~~~ \\
SplineCNN~\cite{SplineCNN2018}  & $99.12\%$~~~ & $99.37\%$~~~ \\
%FMNet~\cite{Litany2017DeepFM}   & $99.33\%$~~~ & $99.54\%$~~~ \\
ACSCNN~\cite{ChebNet}           & $98.98\%$~~~ & $99.64\%$~~~ \\
SPMM-Net                          & $\textbf{99.37\%}$~~~ & $\textbf{99.68\%}$~~~ \\
\hline %inserts single line
\end{tabular}
\label{table:correspondence} % is used to refer this table in the text
\end{table}

\subsection{Further Evaluation}
We conducted further evaluations to assess the effectiveness of the key components in the proposed method on the human body segmentation dataset. The first experiment focused on evaluating the impact of image dense prediction architectures on 3D mesh learning. To this end, we implemented and compared two popular architectures for image segmentation, namely U-Net and HRNetv2. Our results indicate that HRNetv2 outperforms U-Net in 3D mesh learning, similar to the image tasks.

Meanwhile, to evaluate the effectiveness of proposed weighted-average downsample and upsample operations, we tested two architectures with the method proposed in SubdivNet~\cite{hu2021subdivision}. 
We built 4-level mesh sequences with loop connectivity using the method mentioned in the paper, resulting in the same resolution data (16384 faces) as input. We adopted identical network settings and training, while the only differences were the downsampling and upsampling operations. The results show that our method achieves better performance in mesh dense predictions. 

\begin{table}[th]
\centering
\caption{Ablation study on network architectures evaluated on human body segmentation dataset~\cite{Maron2017}.}
\vspace*{3mm}
\begin{tabular}{c c c c} % centered columns (4 columns)
\hline\hline %inserts double horizontal lines
Method & Architecture & Accuracy \\ [0.5ex] % inserts table
\hline % inserts single horizontal line
SubdivNet~\cite{hu2021subdivision} & U-Net~\cite{unet2015}  & $89.5\%$~~~  \\
SPMM-Net & U-Net~\cite{unet2015}  & $90.2\%$~~~  \\
\hline % inserts single horizontal line
SubdivNet~\cite{hu2021subdivision} & HRNet~\cite{HRnet2}   & $90.8\%$~~~  \\
SPMM-Net & HRNet~\cite{HRnet2}  & $91.9\%$~~~    \\
\hline % inserts single horizontal line
\end{tabular}
\label{ablation1}
\end{table}

Table.~\ref{ablation3} compares segmentation networks trained with different input face numbers and downsampling rates. The results show that deeper networks yield higher segmentation accuracy with the same input face number and base face number. We also demonstrate that we can achieve the same base size with different input mesh sizes by adjusting the downsampling rate. No significant difference was observed between the different input sizes.
\begin{table}[th]
\centering % used for centering table
\caption{ Ablation study on decimation algorithm and downsample rate on human body segmentation dataset~\cite{Maron2017}.} % title of Table
\vspace*{3mm}
\begin{tabular}{c c c c c} % centered columns (5 columns)
\hline\hline %inserts double horizontal lines
\#Input faces & rate &\#Paramids faces & Accuracy \\ [0.5ex] % inserts table
%heading
\hline % inserts single horizontal line
16384 & 4 & [16384, 4096, 1024, 256]~~~  & $91.9\%$~~~   \\
16384 & 8 & [16384, 2048, 256]~~~       & $91.1\%$~~~   \\
9216 &  6 & [9216, 1536, 256]~~~        & $91.5\%$~~~   \\
\hline %inserts single line
\end{tabular}
\label{ablation3} % is used to refer this table in the text
\end{table}

%%%%%%%%%%%%%%%%%%%%%%%%%%%%%
%%%%%%%%%%%%%%%%%%%%%%%%%%%%%
\section{Conclusions}
We have presented SPMM-Net, a novel deep learning framework designed for 3D meshes to address dense prediction tasks. Behind SPMM-Net are the construction of a multi-resolution mesh pyramid, incorporating sequential bijective mapping relationships, the design of flexible and robust downsampling/upsampling  and convolution operations for irregular high-resolution meshes, and the design of the multi-resolution network that maintains high-resolution representations throughout the network. 
The key feature of our approach is the bijective inter-surface mapping that redefines meshes, rather than simply reshaping them.
 Extensive experiments have demonstrated that the proposed framework can effortlessly adapt to existing CNN architectures 
and learn robust face features for dense prediction tasks. 

Though our network exhibits flexibility with respect to the connectivity of the input mesh, we require the mesh to be manifold so that the self-parameterization algorithm can be applied. Also, it should be pointed out that the quality of input meshes and the choice of self-parameterization methods impact the pooling and unpooling results.  Another limitation of our method is that it does not support some CNN operations such as strided convolution and dilated convolution. We hope to extend our work to overcome these issues in future.

\section*{References}

\bibliography{mybibfile}

\end{document}